\documentclass[letterpaper, 10 pt, conference]{ieeeconf}  % Comment this line out if you need a4paper

\IEEEoverridecommandlockouts                              % This command is only needed if 
\usepackage[dvipsnames]{xcolor}
\usepackage{amsmath}

\usepackage{booktabs}
\usepackage{hyperref}
\usepackage{caption}
\usepackage{multirow}
\usepackage{url}
\usepackage{todonotes}
\usepackage{algorithm}
\usepackage{algpseudocode}
\usepackage{blindtext}
\usepackage{amsmath}
\usepackage{amsfonts}
\usepackage{bm}
\usepackage{cite}
\usepackage{flushend}
\usepackage[nolist]{acronym}
\usepackage[hang,flushmargin]{footmisc}
\usepackage{graphicx}
\usepackage{subcaption}

\begin{acronym}
\acro{AD}{Autonomous Driving}
\acro{IPP}{Integrated Prediction and Planning}
\acro{IDM}{Intelligent Driver Model}
\end{acronym}

\title{\LARGE \bf 
Perfect Prediction or Plenty of Proposals?\\What Matters Most in Planning for Autonomous Driving
}

\author{Aron Distelzweig$^{1}$, Faris Janjo\v{s}$^{2}$, Oliver Scheel$^{2}$, Sirish Reddy Varra$^{2,3}$, Raghu Rajan$^{1}$, Joschka Boedecker$^{1}$
\thanks{$^1$Department of Computer Science, University of Freiburg, Germany.}%
\thanks{$^2$Bosch Center for Artificial Intelligence, Germany.}%
\thanks{$^3$RWTH Aachen University, Germany.}}

\begin{document}
\maketitle

 \thispagestyle{empty}
\pagestyle{empty}

% ----------------------------- Abstract ------------------------------
\begin{abstract}
Traditionally, prediction and planning in autonomous driving (AD) have been treated as separate, sequential modules. Recently, there has been a growing shift towards tighter integration of these components, known as \ac{IPP}, with the aim of enabling more informed and adaptive decision-making. However, it remains unclear to what extent this integration actually improves planning performance. 
In this work, we investigate the role of prediction in IPP approaches, drawing on the widely adopted Val14 benchmark, which encompasses more common driving scenarios with relatively low interaction complexity, and the interPlan benchmark, which includes highly interactive and out-of-distribution driving situations. Our analysis reveals that even access to perfect future predictions does not lead to better planning outcomes, indicating that current IPP methods often fail to fully exploit future behavior information. Instead, we focus on high-quality proposal generation, while using predictions primarily for collision checks.
We find that many imitation learning-based planners struggle to generate realistic and plausible proposals, performing worse than PDM—a simple lane-following approach. Motivated by this observation, we build on PDM with an enhanced proposal generation method, shifting the emphasis towards producing diverse but realistic and high-quality proposals. This proposal-centric approach significantly outperforms existing methods, especially in out-of-distribution and highly interactive settings, where it sets new state-of-the-art results.
\end{abstract}

% ----------------------------- Intro ---------------------------------
\section{Introduction}
\label{sec:introduction}
Prediction plays a central role in the decision-making pipeline of autonomous driving systems. Accurately forecasting the behavior of surrounding agents is fundamental for understanding the dynamics of traffic scenes and for enabling safe and effective motion planning. In particular, predictions allow the ego vehicle to anticipate possible future developments in the environment and to adapt its maneuvers accordingly.
Many existing systems~\cite{hu2023uniad, zeng2020dsdnet, sadat2020perceivepredictplansafe} follow a modular architecture in which prediction and planning are treated as sequential stages: first, the future trajectories of other agents are predicted, then a separate planner generates a suitable trajectory for the ego vehicle based on these predictions. While this separation simplifies system design and modular evaluation, it limits the planner’s ability to react to and influence the behavior of other agents, especially in dense or highly interactive traffic situations.
\begin{figure}
\captionsetup[subfigure]{skip=0\baselineskip}
    \centering
    \begin{subfigure}{1\columnwidth}
        \includegraphics[width=\columnwidth]{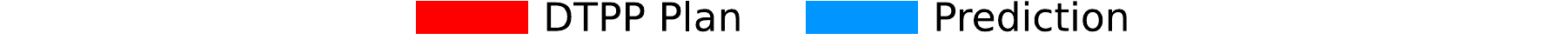}
    \end{subfigure}
    \begin{subfigure}{1\columnwidth}
        \includegraphics[width=\columnwidth]{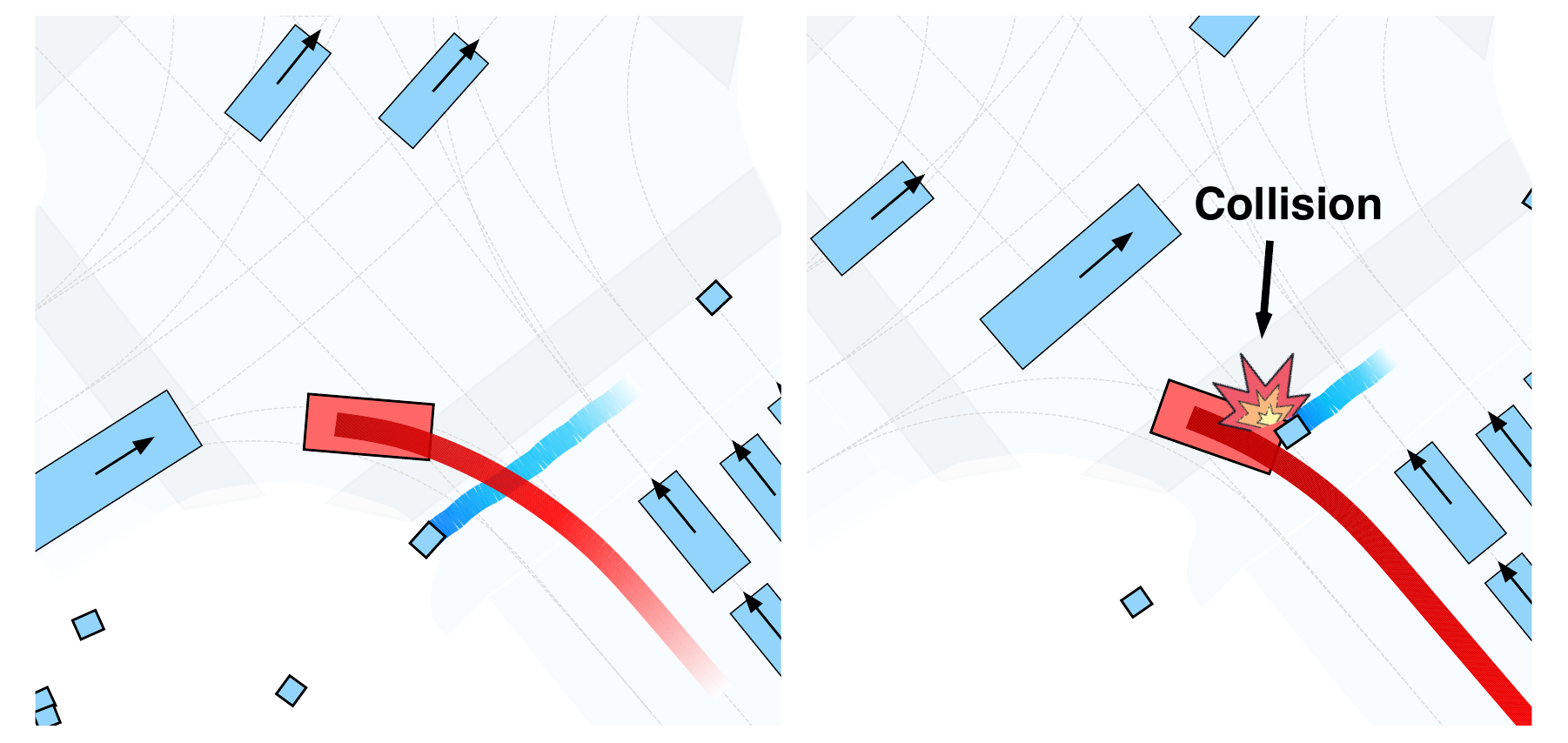}
        \caption{Perfect Prediction}
    \end{subfigure}\\
    \begin{subfigure}{1\columnwidth}
        \includegraphics[width=\columnwidth]{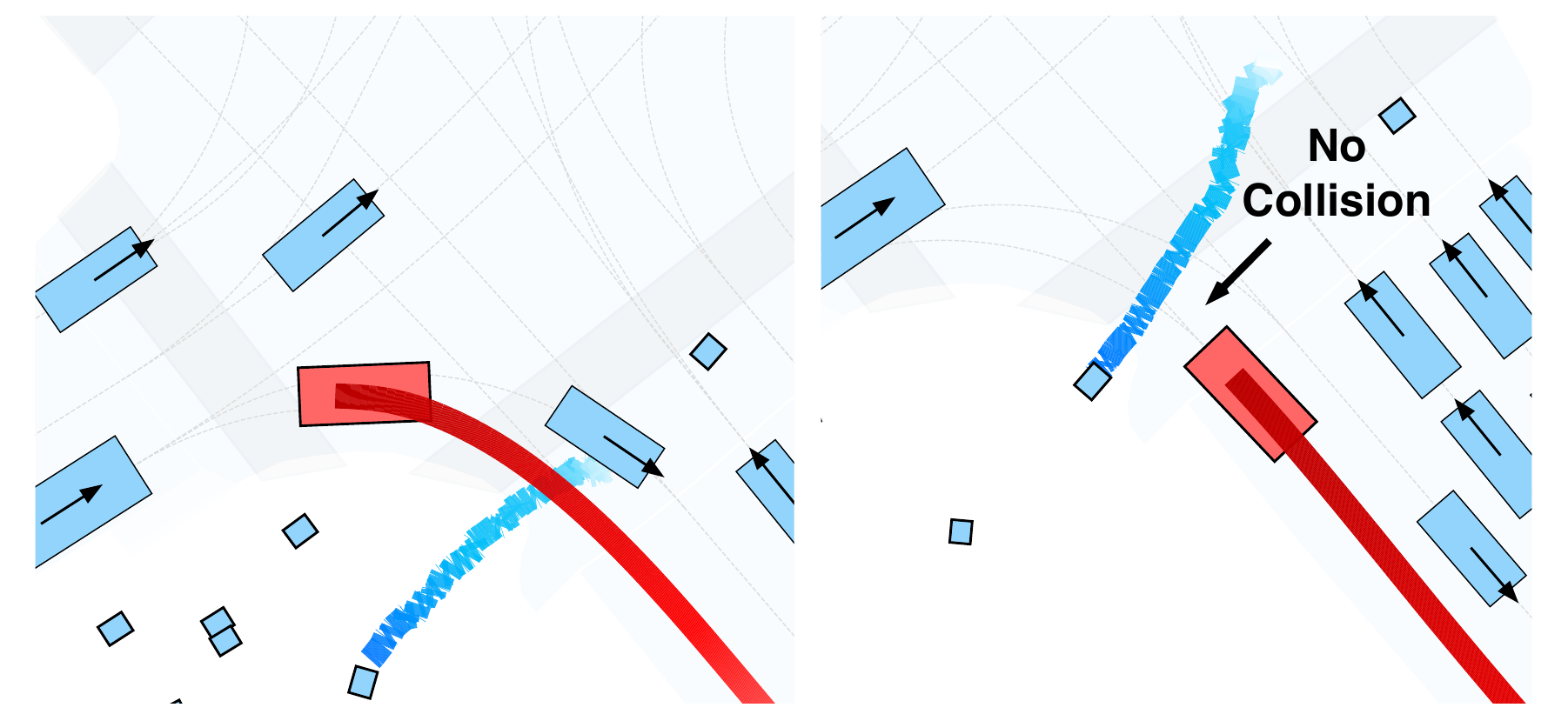}
        \caption{Learned Prediction from DTPP \cite{huang2024dtpp}}
    \end{subfigure}
    \caption{Visualization of a real-world scenario from nuPlan Val14 dataset in reactive simulation. The ego vehicle attempts to cross an intersection while a pedestrian is crossing. Using the state-of-the-art DTPP~\cite{huang2024dtpp} planner with \textbf{(a)} perfect predictions (i.e., the actual responses of agents under the given plan) results in a collision, whereas DTPP with \textbf{(b)} learned predictions successfully avoids the pedestrian. Even when provided with ground-truth future motion, a state-of-the-art planner generates a colliding trajectory. In contrast, it avoids collisions when relying on its own predictions, although these predictions are noisy and inaccurate.}
    \label{fig:dtpp_gt_pred}
\end{figure}
In order to address this limitation, recent approaches~\cite{huang2024dtpp, huang2023gameformer, huang2023dipp} have begun to move towards a tighter integration of prediction and planning. Integrated Prediction and Planning (IPP) methods aim to reason jointly about the evolution of all traffic participants and the ego vehicle, leveraging mutual dependencies to improve decision-making. The underlying hypothesis is that a coupled system can more effectively handle complex interactions, resolve ambiguous situations, and produce more realistic and feasible plans.
While one-stage approaches, whether end-to-end~\cite{hu2023uniad} or mid-to-mid~\cite{bansal2018chauffeurnet}, can be interpreted as forms of IPP, our focus here is on methods with more explicit and tighter integration of prediction and planning.
However, the actual influence of prediction accuracy on planning performance in such integrated systems remains unclear. While more accurate predictions are assumed to translate into better planning outcomes in IPP methods, our results show that this relationship does not hold. Fig. \ref{fig:dtpp_gt_pred} shows an example scenario where the state-of-the-art DTPP Planner~\cite{huang2024dtpp} fails to benefit from perfect predictions, i.e., the true behavior of other agents under a certain plan. Its own predictions are inaccurate and unrealistic, however, they do not yield a collision. We have found similar behavior in other learned IPP approaches. This raises an intriguing question -- \textbf{are planners properly using predictions}?

In order to investigate this issue, we evaluate state-of-the-art IPP approaches under three different conditions: using perfect predictions, learned / rule-based predictions and no predictions. Our findings demonstrate that these methods generally fail to integrate predictions effectively, regardless of prediction accuracy.
Moreover, widely used benchmarks such as the nuPlan datasets Test14~\cite{jcheng2023plantf} and Val14~\cite{dauner2023pdm} do not sufficiently represent the complexity of highly interactive driving scenarios. These benchmarks are biased towards relatively static or straightforward situations, lacking sufficient representation of multi-agent coordination, negotiation, or conflict resolution. Consequently, IPP methods are frequently assessed in settings that do not rigorously challenge their interactive reasoning capabilities.
We complement evaluations on the commonly used Val14 benchmark with evaluations on the more challenging interPlan~\cite{hallgarten2024interplan} benchmark, which emphasizes highly interactive scenarios.

\begin{figure*}[htbp]
    \centering
    \begin{subfigure}[b]{0.4\columnwidth}
        \includegraphics[width=\linewidth]{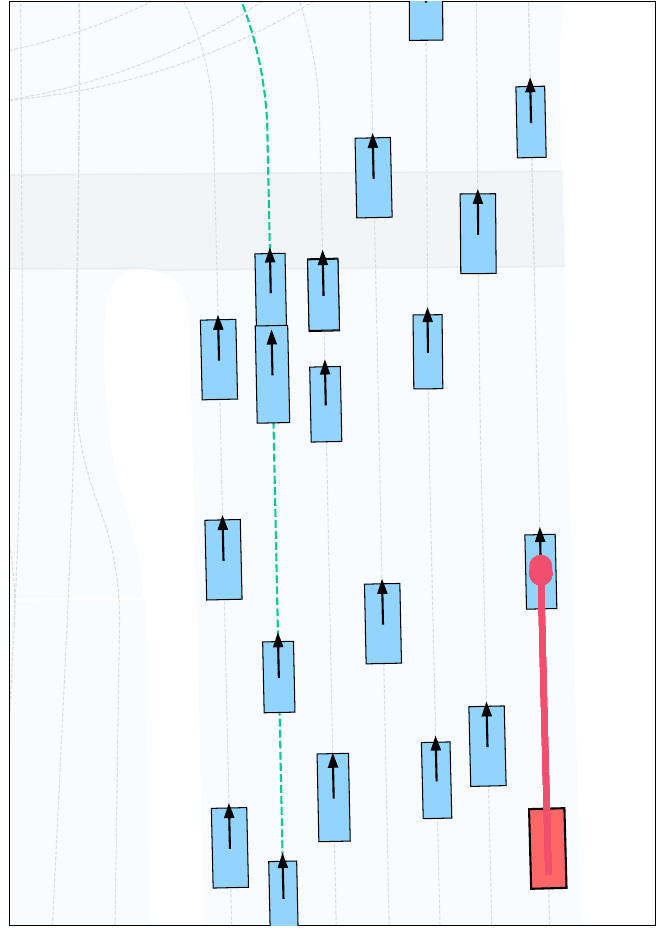}
        \caption{GameFormer\cite{huang2023gameformer}}
        \label{fig:a}
    \end{subfigure}
    \begin{subfigure}[b]{0.4\columnwidth}
        \includegraphics[width=\linewidth]{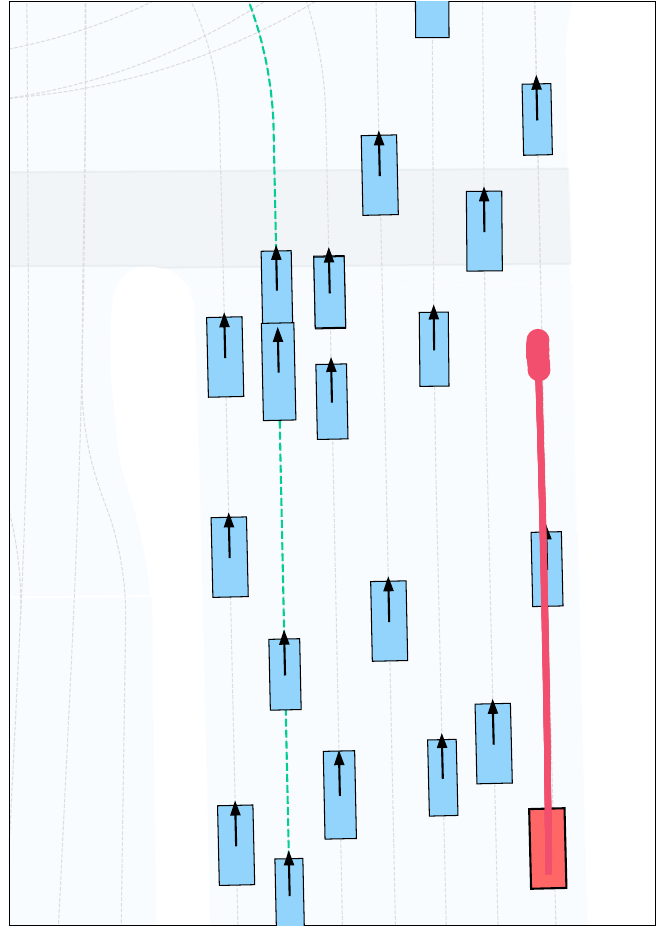}
        \caption{DiffusionPlanner\cite{zheng2025diffusionplanner}}
        \label{fig:b}
    \end{subfigure}
    \begin{subfigure}[b]{0.4\columnwidth}
        \includegraphics[width=\linewidth]{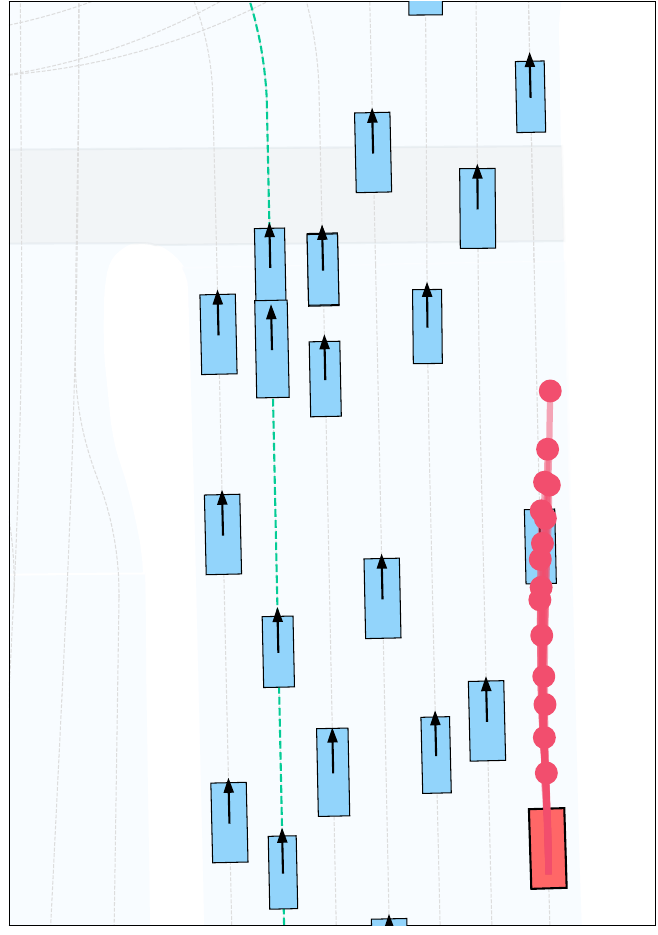}
        \caption{GC-PGP\cite{hallgarten2023gcpgp}}
        \label{fig:c}
    \end{subfigure}
        \begin{subfigure}[b]{0.4\columnwidth}
        \includegraphics[width=\linewidth]{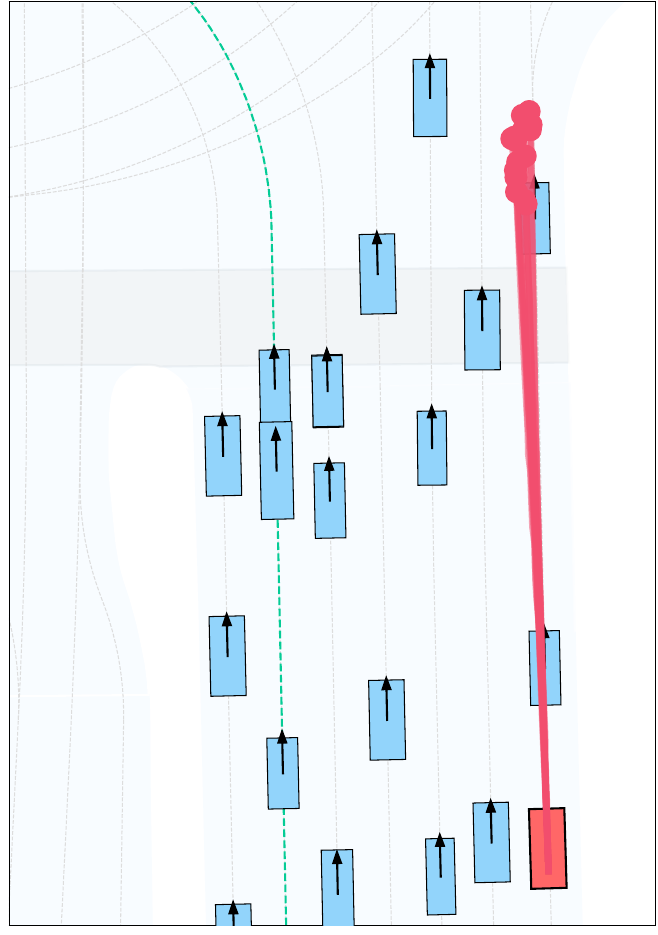}
        \caption{AR GameFormer\cite{huang2023gameformer, philion2024trajeglish}}
        \label{fig:c}
    \end{subfigure}
        \begin{subfigure}[b]{0.4\columnwidth}
        \includegraphics[width=\linewidth]{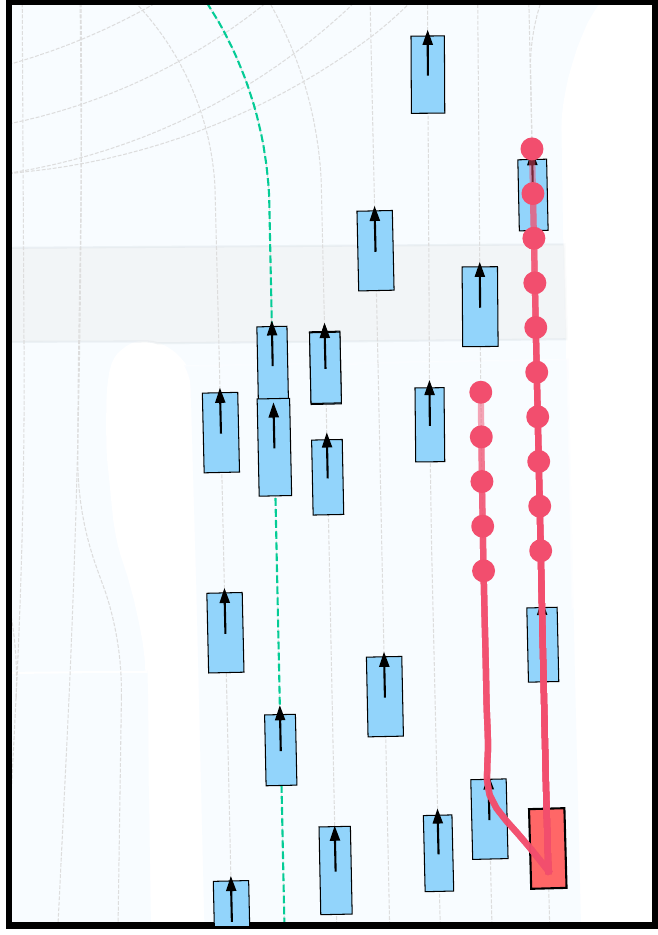}
        \caption{SPDM (Ours)}
        \label{fig:c}
    \end{subfigure}
    \caption{Visualization of $N_p = 15$ ego trajectory proposals generated by various approaches. Each proposal’s endpoint is indicated by a \textcolor{Red}{red} dot. Subfigures (a)–(d) present proposals from learning-based methods, while subfigure (e) illustrates proposals from our rule-based method. In this highly interactive scenario, the ego vehicle must perform multiple consecutive lane changes to reach the designated goal lane, highlighted in \textcolor{Green}{green}. Among the approaches presented, only our proposed method considers a lane change in this scenario.}
    \label{fig:learned_proposals}
\end{figure*}

Additionally, our analysis reveals a critical limitation of many imitation learning-based planning approaches: their tendency to generate trajectories that are neither dynamically feasible nor behaviorally realistic, see Fig. \ref{fig:learned_proposals} for an example. This limitation is evident not only in complex, rare-event interactions requiring advanced anticipatory behaviors as shown in Fig. \ref{fig:learned_proposals}, but also in simpler, more common driving contexts.
Our findings lead us to a shift in emphasis: focus on \textbf{generating strong proposals}, and rely on predictions for collision avoidance during test-time.
To this end, we introduce a rule-based proposal generation method that explicitly models a wide range of realistic maneuvers, including rare and long-tail behaviors. We demonstrate that incorporating such proposals leads to significant performance gains across a range of interactive and out-of-distribution scenarios, outperforming all existing approaches by a large margin.
{In summary, we:}
\begin{enumerate}
    \item Show that perfect predictions in existing \ac{IPP} approaches do not lead to better planning performance.
    \item Demonstrate that current state-of-the-art planners are not able to generate high-quality ego proposals.
    \item Introduce a novel flexible rule-based planner that is able to generate more complex and diverse driving maneuvers than existing planners.
    \item Set new state-of-the-art results on long-tail scenarios in interPlan and outperform all existing approaches by a large margin.
\end{enumerate}

% ----------------------------- Related Work ----------------------------
\section{Related Work}
\label{sec:related_work}
The following sections provide an overview of related work in the areas of prediction, planning, and their integration.
\subsection{Prediction}
Prediction plays a fundamental role in autonomous driving, enabling the vehicle to anticipate the dynamics of its surrounding environment. Broadly, prediction models address two core tasks: (1) context encoding, which involves capturing scene structure and dynamics, and (2) future uncertainty modeling, which accounts for the multi-modal nature of agent behavior. Early prediction models ~\cite{Lee2017Desire, Cui2019Raster, casas2021intentnet} often utilized grid-based scene representations, which, despite their effectiveness, suffered from limited spatial resolution. More recent methods ~\cite{gao2020vectornet, liang2020LGCN} adopt vectorized representations, encoding elements such as lanes, agent trajectories, and road boundaries in a structured format, allowing for richer and more scalable modeling of complex environments. 

To account for the multi-modal nature of agent behavior, prediction models typically generate multiple future trajectories. This can be achieved by generating a set of discrete trajectories~\cite{liang2020LGCN}, conditioning predictions on different map elements (e.g., lanes or goals) ~\cite{Deo2022pgp}, or using latent variable models such as CVAEs ~\cite{salzmann2021trajectron++, Cui2021Lookout, yuan2021agentformer, Faris2024CUAE} or diffusion models ~\cite{jiang2023motiondiffuser}, where different latent samples produce diverse plausible futures. Recent works like ~\cite{philion2024trajeglish, seff2023motionlm} frame trajectory prediction as a next-token language modeling task using discrete representations.

\subsection{Planning}
Planning in autonomous driving involves generating feasible and safe trajectories that navigate towards a goal while accounting for dynamic interactions with the environment. Planning approaches can be broadly categorized into rule-based and learning-based methods. 

Rule-based planners rely on heuristics or physics models, offering interpretability and often safety guarantees. A widely used method is the \ac{IDM}~\cite{Treiber2000IDM}, which models car-following behavior by maintaining a safe distance from the leading vehicle. Extensions like the Predictive Driver Model (PDM)~\cite{dauner2023pdm} incorporate elements of Model Predictive Control (MPC), generating candidate trajectories and selecting the best one based on a cost function. 

Learning-based planners learn driving policies from data and are typically categorized into Imitation Learning (IL) and Reinforcement Learning (RL). IL uses supervised learning to mimic expert behavior. 
%Pioneering work in this area such as ALVINN~\cite{Pomerleau1992Alvinn} demonstrated the use of Convolutional Neural Networks (CNNs)~\cite{lecun1998cnn} for road-following from visual input, and was later extended by end-to-end driving models. 
However, IL suffers from covariate shift, leading to compounding errors~\cite{ross2011IL, hagedorn2025retro}. ChauffeurNet~\cite{bansal2018chauffeurnet} mitigates this issue by injecting perturbations to expert demonstrations, exposing the model to off-distribution states. Recent advances explore transformer-based~\cite{scheel2021urbandriver} and diffusion-based~\cite{yang2024diffusiones, zheng2025diffusionplanner} architectures for motion planning. Goal-Conditioned PGP (GC-PGP)~\cite{hallgarten2023gcpgp} extends the internal graph-based scene representation of the prediction model PGP~\cite{Deo2022pgp}, enabling goal-conditioning for planning. 
RL methods formulate planning as a Markov Decision Process (MDP) and optimize driving policies through interaction. These methods have been applied to lane changing~\cite{Alizadeh2019RLLC}, car following~\cite{zhu2019RLCF}, overtaking~\cite{Meha2018RLOT}, and real-world driving ~\cite{Alex2019DayDrive} tasks. Despite its potential, RL presents challenges including sample inefficiency, safety risks during exploration, and the difficulty of reward function design, although recent advances such as GPUDrive~\cite{kazemkhani2025gpudrive} offer a promising direction to address these limitations.

\subsection{Integrated Prediction and Planning}
Although traditionally treated as separate modules, decoupling prediction and planning can lead to suboptimal behavior, making integration essential for socially-aware decision-making. A central challenge in this process is effectively incorporating predictions into the planning pipeline. According to~\cite{hagedorn2024ipp}, integration strategies can be categorized into five variants: \textit{human leader}, \textit{robot leader}, \textit{implicit}, \textit{joint}, and \textit{co-leader}.
Most existing systems adopt a sequential human leader scheme~\cite{zeng2020dsdnet, hu2023uniad, zeng2021nmp, huang2023dipp}, in which the ego vehicle’s trajectory is conditioned on the predicted motions of other agents. However, this one-way dependency can result in overly conservative behavior~\cite{rhinehart2021cfo}.

To address these limitations, recent research has explored tighter integration through joint reasoning or bidirectional coupling. In joint approaches~\cite{huang2023gameformer, pini2022safepn, zheng2024genad}, the ego and surrounding agents are modeled simultaneously, with their trajectories optimized under a shared global objective. For example, GameFormer~\cite{huang2023gameformer} frames the planning problem as a hierarchical game using a transformer-based architecture that iteratively refines the interactions. However, these methods may be limited by their assumption of shared intent, which may not hold in real-world traffic.
In contrast, bidirectional integration methods such as co-leader approaches model the mutual influence between the ego’s plan and predicted behaviors of others, without assuming a shared objective, which makes them more flexible in capturing interactive behaviors. ~\cite{rhinehart2021cfo} proposes a method that employs a learned behavioral model to stochastically forecast multi-agent trajectories, combined with a planning objective to compute contingency plans. In contrast,~\cite{Chen2023tpp, huang2024dtpp} formulate planning as a discrete MDP with two trees, representing ego and other agents’ potential behaviors, solved via dynamic programming.

End-to-End approaches~\cite{hu2023uniad, jiang2023VAD} fall into the broader category of IPP, however, our focus is on advanced integration methods, which can often be embedded within End-to-End frameworks, so the methods we study can be seen as specific building blocks of the broader End-to-End paradigm.
\subsection{Benchmarks}
In recent years, several large-scale datasets and simulation frameworks have been developed to standardize the evaluation of autonomous driving systems. Some benchmarks focus primarily on open-loop prediction~\cite{Caesar2019nuscenes, Wilson2021Argoverse2, ettinger2021womd}, while others support closed-loop planning evaluation~\cite{Dosovitskiy2017carla, caesar2022nuplan}. Given the interdependence of prediction and planning, it is increasingly recognized that \ac{IPP} systems should be evaluated as a whole.
nuPlan~\cite{caesar2022nuplan} is one of the most established and relevant frameworks for planning. It provides a simulation framework built on real-world driving logs and supports open-loop and closed-loop evaluation settings. In the non-reactive setting, the surrounding agents follow recorded trajectories from the dataset. In the reactive setting, their behavior is modeled using Intelligent Driver Model (IDM)~\cite{Treiber2000IDM}, allowing for limited interaction with the ego. The Val14 benchmark is the standardized evaluation split for nuPlan, but it primarily consists of straightforward, low-interaction scenarios that can often be solved with basic lane-following behavior. To address this limitation, interPlan~\cite{hallgarten2024interplan} augments the nuPlan dataset by introducing rare long-tail scenarios through manual scene modification.

% ----------------------------- APPROACH ------------------------------
\section{METHODOLOGY}
\label{sec:approach}
To systematically investigate Integrated Prediction and Planning (IPP) approaches, we select a set of representative and widely recognized methods from the literature. 
These include: Differentiable Tree Policy Planning (DTPP)~\cite{huang2024dtpp}, a tree-based co-leader strategy; GameFormer~\cite{huang2023gameformer}, which jointly models multi-agent interactions through a game-theoretic framework; and Differentiable Integrated Prediction and Planning (DIPP)~\cite{huang2023dipp}, a non-linear optimization-based method that jointly incorporates prediction into the planning process.

Section \ref{subsec:ipp_methods} provides an overview of how each selected approach integrates prediction and planning. For in-depth methodological details, we refer the reader to the respective original work. Section \ref{subsec:pred_mods} presents the methodology employed to generate perfect prediction models across different simulation settings. Finally, Section \ref{subsec:spdm} presents our proposed proposal generation method.

\subsection{Integrated Prediction and Planning Approaches}
\label{subsec:ipp_methods}
The following section provides a brief overview of the three distinct IPP approaches: DTPP, GameFormer, and DIPP with a particular emphasis on how each method integrates prediction and planning.

\textbf{Differentiable Tree Policy Planning (DTPP)}~\cite{huang2024dtpp}
constructs a trajectory tree through iterative node expansion, guided by learned ego-conditioned prediction and cost models, until a predefined maximum number of expansion stages, $N_l$, is reached. At each stage $k$, every branch of the ego trajectory tree is expanded by a rule-based proposal mechanism. The ego-conditioned prediction model then forecasts the future trajectories of surrounding agents conditioned on these ego proposals. 
The prediction model follows a Transformer encoder--decoder architecture. The encoder maps agent histories 
$A \in \mathbb{R}^{N_a \times T_h \times d_a}$, where $N_a$ is the number of surrounding agents, $T_h$ the history length, and $d_a$ the agent attributes, together with map features $M \in \mathbb{R}^{N_m \times N_p \times d_m}$, where $N_m$ is the number of map elements, $N_p$ the number of waypoints, and $d_m$ the map attributes, into a scene context $C = \mathbf{Enc}(A, M)$. The decoder then produces ego-conditioned predictions $Y^{(k)} = \mathbf{Dec}(C, E^{(k)}) \in \mathbb{R}^{B \times N_a \times T_f^k \times 2}$ conditioned on the scene context and the ego trajectory tree $E^{(k)} \in \mathbb{R}^{B \times T_f^k \times 3}$, where $B$ is the number of branches and $T_f^k$ the future length at stage $k$. Based on these interaction-aware predictions, a learned cost function  evaluates and ranks ego trajectory candidates $s = c(E^{(k)}, Y^{(k)})$, selecting the most promising ones for further expansion. The cost function is optimized via Inverse Reinforcement Learning (IRL). In practice, the number of expansion stages is set to $N_l = 2$.

\textbf{GameFormer}~\cite{huang2023gameformer}
unifies prediction and planning within a game-theoretic framework.  Agent histories $A$ and map features $M$ are first encoded into a scene context representation $C = \mathbf{Enc}(A, M)$ using a Transformer encoder. On top of this context, $N_l$ Transformer-based decoders model multi-agent interactions across different reasoning levels. 
At level-$0$, the decoder jointly predicts the future trajectories of all agents, including the ego vehicle, 
$Y_e^{(0)} = \mathbf{Dec}^{(0)}(C) \in \mathbb{R}^{N_a \times T_f \times 2}$, where modality embeddings and agent histories serve as queries and the scene context $C$ provides the keys and values. We denote $Y_e^{(k)} \in \mathbb{R}^{(N_a+1) \times T_f \times 2}$ as the predictions at level-$k$ for all agents including the ego agent.
At level-$k$ ($k \geq 1$), the decoder refines predictions by incorporating the trajectories from the previous level, $Y_e^{(k)} = \mathbf{Dec}^{(k)}(C, Y_e^{(k-1)}).$
Through self-attention, each agent attends to the trajectories of all other agents while masking its own, thereby modeling interactive responses. Predictions are iteratively refined over $N_l = 3$ levels, resulting in a hierarchy of interaction-aware future trajectories.

\textbf{Differentiable Integrated Prediction and Planning (DIPP)}~\cite{huang2023dipp}
employs a neural network architecture to jointly predict the future trajectories of surrounding agents, denoted as $Y_{e,i\geq 1}$, and an initial plan for the ego vehicle, denoted as $Y_{e,i=0}$. At its core lies a differentiable optimization module that refines the initial ego plan by accounting for the predicted behaviors of other agents and optimizing multiple objectives, including compliance with speed limits, passenger comfort, adherence to traffic rules, and safety, measured in terms of collision avoidance and maintaining appropriate distances to other vehicles. The optimization problem is formulated as
$u^* = \underset{u}{\arg\min}\; \tfrac{1}{2} \sum_i \left\| \omega_i\, c_i(u, Y_{e,i\geq 1}) \right\|^2$, where $u = \{u_1, u_2, \dots, u_T\}$ are the control inputs to be optimized, $c_i$ are cost terms, and $\omega_i$ are their corresponding weights. The predicted initial plan $Y_{e,i=0}$ serves as initialization for the control inputs.  
In contrast to the original work, we employ GameFormer with $N_l=0$ levels to predict the surrounding agents’ future trajectories as well as the ego’s initial plan. Furthermore, we integrate GameFormer with $N_l=3$ levels into the DIPP framework, denoted as \textit{GameFormer + DIPP}.

\subsection{Prediction Modalities in IPP Methods}
\label{subsec:pred_mods}
To examine the impact of predictions on planning performance, we systematically vary the type of predictions used in the selected IPP methods.
Let $Y \in \mathbb{R}^{N_a \times T_f \times 2}$ be the $x, y$ positions of $N_a$ neighboring agents for $T_f$ future timesteps. In our experiments, we set the number of agents, $N_a=10$, corresponding to the nearest agents to the ego vehicle.
We consider the following conditions:
\subsubsection{Learned / Rule-based Predictions}
Set $Y = \hat{Y}$, where $\hat{Y} \in \mathbb{R}^{N_a \times T_f \times 2}$ are predictions generated by a learned or rule-based prediction module.
\subsubsection{Perfect Predictions}
The actual future behavior of other agents is provided, i.e., $Y = Y^*$, where $Y^* \in \mathbb{R}^{N_a \times T_f \times 2}$ denotes the ground-truth trajectories of surrounding agents. In the non-reactive closed-loop and open-loop setting, $Y^*$ corresponds to recorded future trajectories available in the dataset. For reactive simulations, we utilize the Intelligent Driver Model (IDM)~\cite{Treiber2000IDM} to generate behavior that responds to the actions of the ego vehicle. Since pedestrians are not simulated via IDM in nuPlan, their recorded future behavior is used instead, even in reactive settings.

\subsubsection{No Predictions}
no predictions are made about the future behavior of other agents; we mask out the futures of all agents by $Y=\mathbf{0}$.

We implement the conditions above in a manner tailored to each model. In DTPP, predictions are replaced at every stage whereas in GameFormer they are replaced in each decoder. In DIPP, predictions are replaced before the refinement procedure.

\subsection{Spline-Fitting PDM (SPDM)}
\label{subsec:spdm}
Our method for generating improved ego proposals builds upon the PDM-Closed approach and the proposal generation strategy from~\cite{huang2024dtpp}. PDM-Closed generates IDM-based proposals along the centerline with varying target velocities, while assuming constant velocity for the other agents. It then selects the optimal proposal using a scoring function closely aligned with the nuPlan metric score.
Our approach comprises two stages.
In the first stage, following the PDM methodology, we generate IDM proposals along the current centerline by varying the target velocity across five values, $\{20, 40, 60, 80, 100\}\%$ of the current speed limit, and applying three lateral offsets, $\{-1 , 0, +1\}\text{m}$.

Our spline-fitting extension expands the set by including trajectories that are not constrained to the current centerline. To this end, we extract all lanes along the goal-directed route, referred to as route lanes, and generate additional proposals for each route lane using the same target speed variations as in the first stage.
We model the velocity profile over time between the current time and the time horizon, $T_f$. Let the initial ego state be defined as $s(0) = [x_0, y_0, v_0, a_0, \theta_0, l_0]$, where the components denote position coordinates, velocity, acceleration, heading angle, and lane position, respectively. For each target velocity, we fit a cubic polynomial to describe the velocity profile over time $t$:
\begin{align}
v(t) &= c_0 + c_1 t + c_2 t^2 + c_3 t^3, \\
v(0) &= v_0,\quad v(T_f) = v_{\text{target}}, \\
\dot{v}(0) &= a_0,\quad \dot{v}(T_f) = 0.
\end{align}
Here, $(c_0, c_1, c_2, c_3)$ are polynomial coefficients.
This formulation ensures smooth transitions in both velocity and acceleration, thereby enabling the generation of realistic and diverse ego trajectories that are not restricted to a single lane. The proposals generated from both stages are merged into a unified candidate set. Following the PDM procedure, control actions are then iteratively computed using a Linear Quadratic Regulator (LQR) controller, and the ego state is propagated using a kinematic bicycle model \cite{polack2017bicycle}. All candidate proposals are then evaluated using the PDM scoring function inspired by the nuPlan metrics. This function considers factors such as time-to-collision, collisions, progress, comfort, drivable area adherence, and compliance with driving direction. The proposal with the highest score is chosen as the final plan. Overall, our SPDM method is lightweight, produces realistic and kinematically feasible trajectories, and can be seamlessly integrated into existing motion generation pipelines.

% ----------------------------- RESULTS ----------------------------
\section{Experiments}
\label{sec:results}

\begin{table}
\resizebox{\columnwidth}{!}{
\centering
\begin{tabular}{llcccc}
\toprule
\multirow{2}{*}{Model} & \multirow{2}{*}{Predictions} & \multicolumn{3}{c}{Val14} & \multicolumn{1}{c}{interPlan} \\
\cmidrule(lr){3-5} \cmidrule(lr){6-6}
& & NR$\uparrow$ & R$\uparrow$ & \multicolumn{1}{c}{OL$\uparrow$} & R$\uparrow$ \\
\midrule
\multirow{3}{*}{PDM-Closed\cite{dauner2023pdm}} & Constant Velocity & 92.84 & 92.28 & 42.52 & 41.88 \\
 & Perfect & \textbf{93.80} & N/A & \textbf{46.96} & N/A  \\
 & None & 41.07 & 40.33 & 15.74  & 30.63 \\
\midrule
\multirow{3}{*}{DTPP\cite{huang2024dtpp}} & Learned & \textbf{64.54} & \textbf{68.23}  & \textbf{67.05}  & \textbf{37.27} \\ % l2 in OL: 0.71
 & Perfect & 53.65 & 64.04 & 32.52   & 25.12 \\ % l2 in OL: 0.47
 & None & 51.09 & 54.81 & 56.49  & 21.72 \\
\midrule
\multirow{3}{*}{GameFormer\cite{huang2023gameformer}} & Learned & \textbf{34.99} & \textbf{34.98}  & \textbf{76.33}  & \textbf{4.10} \\ % l2 in OL: 0.71
 & Perfect & 33.92 & 34.82 & 76.31 & 3.28 \\ % l2 in OL: 0.47
 & None & 30.24 & 26.71 & 76.26  & 3.28 \\
\midrule
\multirow{3}{*}{GameFormer\cite{huang2023gameformer} + DIPP\cite{huang2023dipp}} & Learned & 79.83 & 80.95  & 80.38  & \textbf{21.26} \\
 & Perfect & \textbf{81.04} & \textbf{81.15} & \textbf{81.01} & 19.00 \\ 
 & None & 58.14 & 51.14 & 80.73  & 13.19 \\
\midrule
\multirow{3}{*}{DIPP\cite{huang2023dipp}} & Learned & 73.35 & 77.65 & 37.21 & \textbf{21.25} \\
 & Perfect & \textbf{74.19} & \textbf{79.02} & \textbf{37.67}  & 12.90 \\
 & None & 49.58 & 56.86 & 37.25  & 13.07 \\
\bottomrule
\end{tabular}
}
\caption{Performance of various integrated prediction and planning approaches on the Val14 and interPlan benchmarks, evaluated under different prediction sources: learned/rule-based, perfect predictions, or no predictions. NR/R: non-reactive/reactive simulation and OL: open-loop simulation.}
\label{tab:preds_influence}
\end{table}

\begin{figure}[t]
\centering
\begin{center}
\includegraphics[width=\columnwidth]{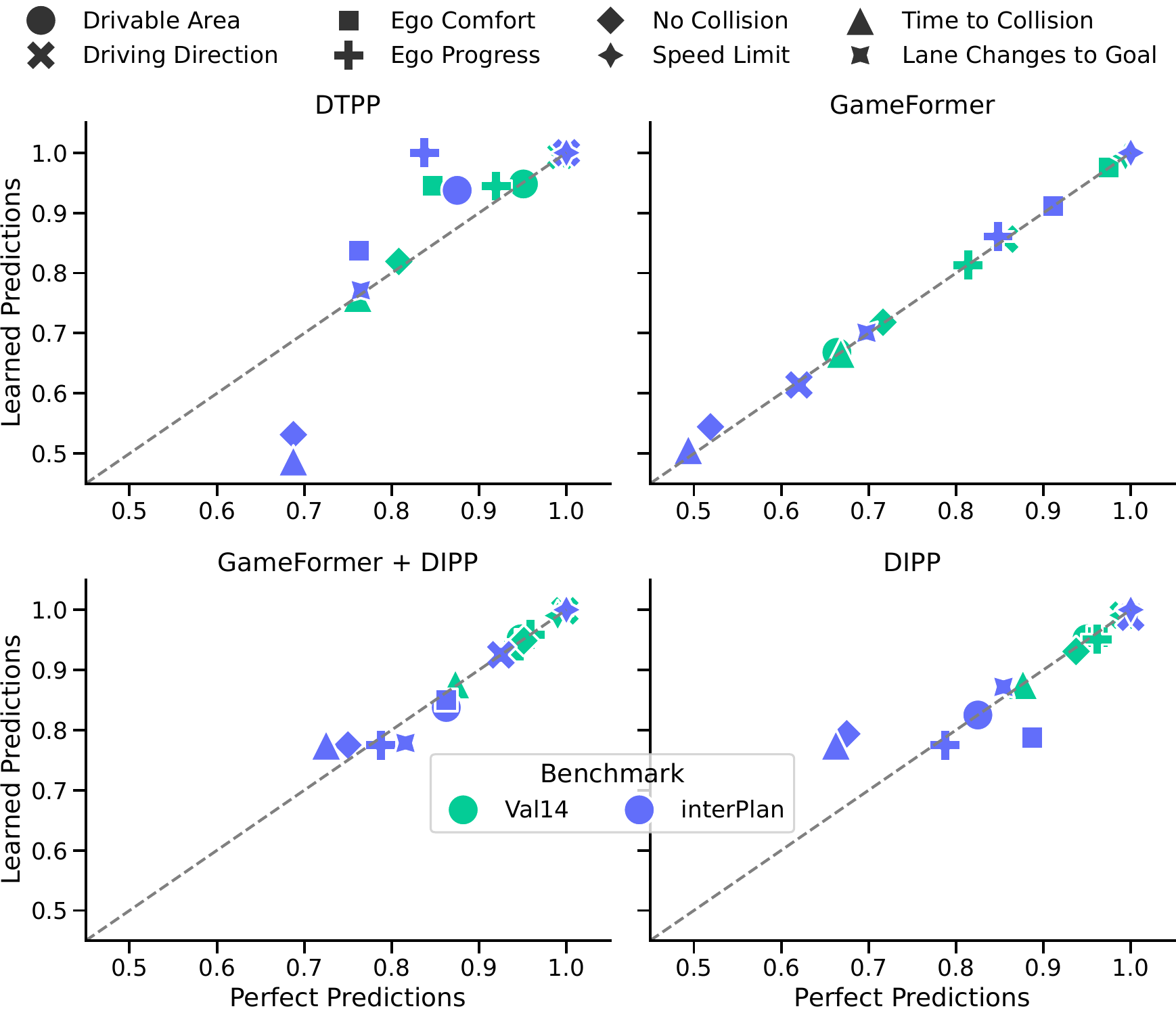}
\end{center}
\caption{Comparative analysis of subscores between learned and perfect predictions when utilized within DTPP, GameFormer, GameFormer + DIPP, and DIPP across the Val14 and interPlan benchmark in reactive simulation. Markers in the upper triangle denote better performance with learned predictions, whereas markers in the lower triangle denote better performance with perfect predictions. Subscore ‘Lane Changes to Goal’ is exclusively available in the interPlan benchmark.}
\label{fig:learned_vs_perfect}
\end{figure}

% ##################### SUBSECTION: TRAINING AND EVALUATION #####################
\subsection{Evaluation}
We employ the nuPlan framework~\cite{caesar2022nuplan}, a closed-loop simulator based on real-world driving data. We adopt the widely used Val14 benchmark~\cite{dauner2023pdm}, comprising of 1,118 scenarios drawn from the nuPlan validation set. Additionally, we consider the recently introduced out-of-distribution interPlan benchmark~\cite{hallgarten2024interplan}, which consists of 80 complex driving scenarios. These include maneuvering around parked vehicles, responding to jaywalkers, and executing lane changes in high traffic density.
For performance assessment, we use the closed-loop score~\cite{karnchanachari2024nuplan}, a weighted composite metric that combines, among others, progress, collision, time-to-collision, speed limit compliance, and comfort into a single 0–100 scale. InterPlan extends this score by including the number of lane changes as an additional evaluation component. Both benchmarks simulate background traffic using IDM in reactive closed-loop setting.

\begin{table}[t]
\resizebox{\columnwidth}{!}{
\centering
\begin{tabular}{llcccc}
\toprule
\multirow{2}{*}{Type} & \multirow{2}{*}{Proposal Generator} & \multirow{2}{*}{$N_p$} & \multicolumn{2}{c}{Val14} & \multicolumn{1}{c}{interPlan} \\
\cmidrule(lr){4-5} \cmidrule(lr){6-6}
& & & NR$\uparrow$ & R$\uparrow$ & R$\uparrow$ \\
\midrule
\multirow{1}{*}{Rule-based} 
    & PDM-Closed~\cite{dauner2023pdm} & 15 & \textbf{92.84} & \textbf{92.28} & 41.88 \\
\midrule
\multirow{3}{*}{Diffusion} 
    & Diffusion-ES*~\cite{yang2024diffusiones} & 15 & 75.98 & 79.92 & \textbf{57.41} \\
    & Diffusion-ES~\cite{yang2024diffusiones} & 15 & 72.77 & 75.80 & 35.18 \\
    & DiffusionPlanner~\cite{zheng2025diffusionplanner} & 15 & 91.26 & 91.13 & 25.76 \\
\midrule
\multirow{2}{*}{NAR Transformer} 
    & GameFormer\cite{huang2023gameformer} + DIPP\cite{huang2023dipp} & 15 & 82.98 & 83.39 & 19.09 \\
    & GameFormer\cite{huang2023gameformer} & 15 & 42.95 & 40.30 & 3.95 \\
\midrule
AR Transformer & AR GameFormer~\cite{huang2023gameformer, philion2024trajeglish} & 15 & 43.46 & 44.22 & 1.02 \\
\midrule
Graph-based & GC-PGP~\cite{hallgarten2023gcpgp} & 15 & 58.40 & 63.82 & 14.55 \\
\bottomrule
\end{tabular}
}
\caption{PDM planner performance on the Val14 and interPlan benchmarks across a diverse set of proposal generators, with the best proposal selected using the PDM scoring function. Our comparison spans a broad range of model families, including rule-based, diffusion, transformer-based, and graph-based approaches. Diffusion-ES* uses PDM proposals as initial set instead of learned proposals. NR/R: non-reactive/reactive simulation}
\label{tab:proposal_gens}
\end{table}

% ##################### SUBSECTION: PREFECT PREDICTIONS DO NOT IMPROVE PLANNING #####################
\subsection{Perfect Predictions Do Not Improve Planning}
\label{sec:perfect_preds}
To assess whether IPP methods utilize predictions in a meaningful manner, we systematically modified the predictions within these approaches, as outlined in Section \ref{subsec:pred_mods}. The outcomes of this analysis are presented in Table \ref{tab:preds_influence}.
PDM-Closed follows a sequential integration of predictions and, therefore, cannot be evaluated with perfect predictions in reactive closed-loop simulations. This constitutes a clear limitation, as other agents dynamically adapt their behavior in response to the ego agent, rendering fixed predictions inherently suboptimal. Nonetheless, we include PDM-Closed as a baseline. The results reveal a consistent trend: all evaluated IPP methods fail to effectively leverage predictions. Even when provided with perfect predictions, none of the approaches demonstrate a significant improvement in planning performance. Notably, the co-leader method DTPP exhibits a decline in performance across all simulation settings compared to when including learned predictions. GameFormer and DIPP exhibit comparable performance when utilizing perfect predictions compared to their performance with learned predictions on the Val14 benchmark. However, on the interPlan benchmark, both methods show a reduction in performance when provided with perfect predictions.

Across all methods, a complete removal of predictions results in a decrease in performance, indicating that knowledge of future scene dynamics generally contributes to improved planning outcomes. However, it is counterintuitive that the use of perfect predictions does not yield a significant performance gain.
One could argue that this results from the distributional shift between real-world driving and simulation. However, we also observe no improvement when using ground-truth data in the non-reactive setting, indicating that even with in-distribution data, perfect predictions do not yield performance gains.
In order to assess whether access to perfect predictions yields improvements in specific subscores, we decompose the final evaluation scores into their constituent subscores, as shown in Figure~\ref{fig:learned_vs_perfect}. The analysis indicates that only DTPP exhibits a significant improvement in both collision rate and time-to-collision on the interPlan benchmark, while none of the other approaches show any statistically significant improvement. Interestingly, DTPP exhibits only a marginal performance drop on the Val14 benchmark. This observation is consistent with the fact that Val14 scenarios lack high interactivity.
GameFormer and GameFormer + DIPP demonstrate comparable performance under both learned and perfect prediction settings. Notably, DIPP shows a degradation in collision-related metrics when using perfect predictions on the interPlan benchmark. These findings suggest that, among the evaluated methods, only DTPP exhibits a limited ability to leverage perfect predictions for reducing collisions. However, the overall limited improvement underscores a substantial gap in the effective utilization of predictions.
Despite having access to the true behavior of other agents, these methods continue to exhibit critical failure cases, thereby revealing a significant limitation in their capacity to exploit predictions.

% ##################### SUBSECTION: LEARNING-BASED APPROACHES ARE BAD PLANNERS #####################
\subsection{Learning-based Approaches Are Ineffective Proposal Generators}
\label{sec:bad_planners}

\begin{figure}[t]
\centering
\begin{center}
\includegraphics[width=\columnwidth]{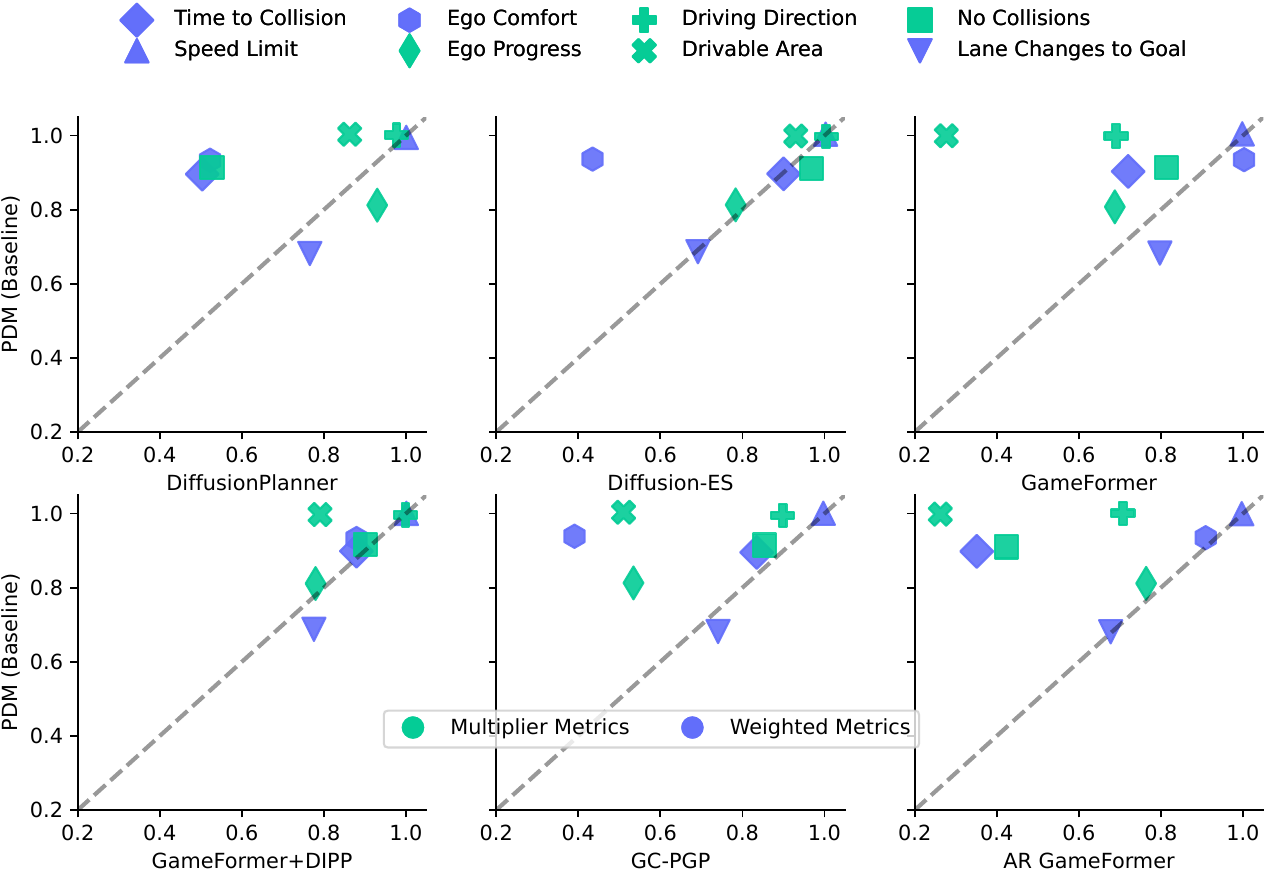}
\end{center}
\caption{Subscore performance comparison between learned-based proposal generators and the rule-based PDM-Closed on the interPlan benchmark in reactive simulation. Each point represents a specific subscore, with the y-axis indicating PDM-Closed performance and the x-axis showing performance of the corresponding learned-based generator. For the learned-based methods, the best proposal is selected using the PDM scoring function.}
\label{fig:spdm_vs_learned}
\end{figure}
%\begin{figure}[t]
%\centering
%\begin{center}
%\includegraphics[width=.6\columnwidth]{figures/fig5.pdf}
%\end{center}
%\caption{Performance of PDM across varying numbers of proposals on the Val14 and interPlan benchmarks. NR/R: non-reactive/reactive simulation}
%\label{fig:proposals}
%\end{figure}

\begin{figure}
\captionsetup[subfigure]{skip=0\baselineskip}
    \centering
    \begin{subfigure}{.49\columnwidth}
        \includegraphics[width=\columnwidth]{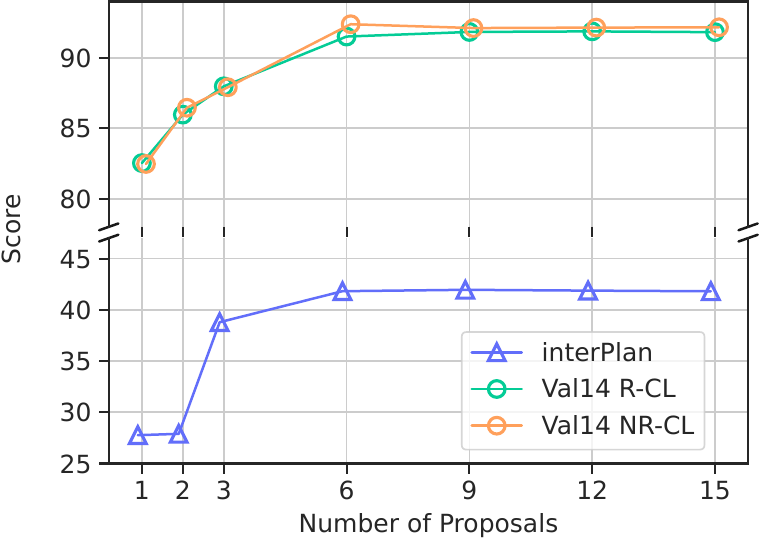}
        \caption{}
        \label{fig:proposals}
    \end{subfigure}
    \begin{subfigure}{.49\columnwidth}
        \includegraphics[width=\columnwidth]{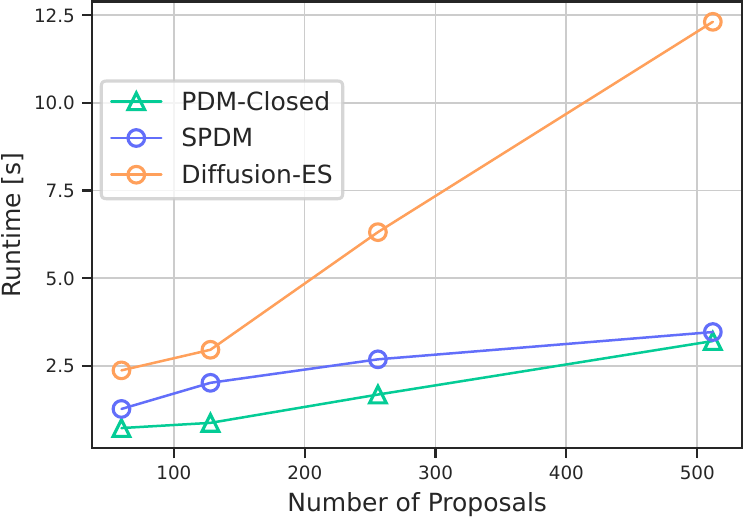}
        \caption{}
        \label{fig:runtime}
    \end{subfigure}\\
    \caption{Figure (a) shows performance of PDM across varying numbers of proposals on the Val14 and interPlan benchmarks. NR/R: non-reactive/reactive simulation. Figure (b) shows runtime analysis across varying numbers of proposals, averaged over 10 distinct scenarios.}
\end{figure}

A critical component of IPP methods is the plan generation process, which can be broadly categorized into rule-based and learning-based approaches. Rule-based methods typically generate multiple candidate plans from which the most suitable one is selected. In contrast, learning-based approaches aim to directly produce an optimal plan in a single inference step, offering greater adaptability and efficiency. 

We aim to investigate whether learning-based approaches can be used as proposal generators. To this end, we compare learning-based planners with the rule-based PDM planner.
However, our observations indicate that existing imitation learning-based planners often exhibit significant limitations in planning efficacy. Specifically, they consistently fail to produce trajectories that yield high-quality driving performance. To empirically support this claim, we selected a set of state-of-the-art learning-based approaches from diverse model families and used each to generate multiple trajectory proposals. We then evaluate these proposals using a scoring function closely aligned with the nuPlan metric and select the highest-scoring proposal as the final plan.
Table \ref{tab:proposal_gens} presents the results for $N_p = 15$ proposals generated per method. 
For GameFormer~\cite{huang2023gameformer} and GC-PGP~\cite{hallgarten2023gcpgp}, we set the number of modes to $m = 15$ and retrain the models accordingly. In the case of DiffusionPlanner~\cite{zheng2025diffusionplanner}, we set the sampling temperature to $t = 1.5$ to encourage greater diversity in the generated trajectories. Diffusion-ES~\cite{yang2024diffusiones} combines PDM-generated proposals with additional learned ones. However, to ensure fair comparison across methods, we also evaluate a variant that excludes the rule-based PDM proposals from the initial candidate set. The autoregressive transformer-based encoder–decoder architecture constitutes an autoregressive adaptation of the GameFormer architecture. For tokenization, we follow the approach proposed in~\cite{philion2024trajeglish}, and generate diverse proposals using nucleus sampling~\cite{holtzman2020nucleus}.
The findings indicate that proposals from PDM significantly outperform most learning-based approaches, particularly on the interPlan benchmark, where the performance gap is most pronounced. An exception is Diffusion-ES*, which achieves strong results, likely due to its initialization with PDM proposals. Notably, all evaluated learning-based planners were trained using motion forecasting objectives via imitation learning.

In Figure \ref{fig:spdm_vs_learned}, we present a detailed comparison of subscore metrics between PDM and the highest-quality proposal among the $N_p = 15$ generated by each respective learning-based method. Across most subscore categories, PDM significantly outperforms the learning-based baselines.
Prior work ~\cite{dauner2023pdm} has already highlighted a fundamental misalignment between the objectives of motion forecasting and planning. Our comprehensive evaluation reveals that purely imitation learning-based approaches consistently fail to produce good plans, not only in rare or out-of-distribution scenarios as in the interPlan benchmark, but also in common and relatively simple situations as in the Val14 benchmark.

\begin{table}
\centering
\begin{tabular}{lcccc}
\toprule
\multirow{2}{*}{Proposal Generator} & \multirow{2}{*}{$N_p$} & \multicolumn{2}{c}{Val14} & \multicolumn{1}{c}{interPlan} \\
\cmidrule(lr){3-4} \cmidrule(lr){5-5}
& & NR$\uparrow$ & R$\uparrow$ & R$\uparrow$ \\
\midrule
PDM-Closed\cite{dauner2023pdm} & 60 & 91.21 & 91.26 & 42.96 \\
\midrule
\multirow{6}{*}{SPDM (Ours)} & 15 & \textbf{92.84} & \textbf{92.28} & 42.00 \\
& 20 & 92.73 & 92.21 & 49.72 \\
& 25 & 92.45 & 92.04 &  52.77 \\
& 30 & 92.11 & 91.91 & 59.07 \\
& 45 & 91.75 & 91.85 & 62.37 \\
& 60 & 91.49 & 91.60 & \textbf{63.66} \\
\bottomrule
\end{tabular}
\caption{Performance analysis of the proposed spline-fitting PDM method across different numbers of proposals $N_p$ on the Val14 and interPlan benchmarks. NR/R: non-reactive/reactive simulation.}
\label{tab:proposals_spdm_pdm}
\end{table}
% ##################### SUBSECTION: HIGH QUALITY PROPOSALS IMPROVE PLANNING #####################
\subsection{High Quality Proposals Improve Planning}
Building on the findings presented in Section \ref{sec:perfect_preds}, we aim to further investigate the underlying causes of the observed results. While some performance limitations may arise from the characteristics of individual planning approaches, such as the learned cost function in DTPP, in our analysis, we intentionally avoid attributing these limitations to specific methods. Instead, we hypothesize the existence of a shared structural bottleneck across planning approaches. Specifically, we question whether improvements in predictions can meaningfully enhance planning performance, or whether advancements must instead stem from the plan generation process itself.
As demonstrated in Section \ref{sec:bad_planners}, learning-based methods have shown limited effectiveness in generating high-quality plans. Consequently, we shift our focus to rule-based planning strategies. In an initial experiment, we evaluate whether the PDM proposals can contribute to improved driving performance.

Figure \ref{fig:proposals} illustrates that decreasing the number of proposals (originally set to $N_p = 15$) results in a performance drop when fewer than $N_p = 6$ proposals are used. Notably, even with as few as $N_p = 3$, the performance still surpasses that of most learning-based planners on both the Val14 and interPlan benchmarks. Furthermore, increasing $N_p$ beyond the threshold value of $6$ does not yield additional performance gains, indicating a saturation point for the benefit of simple trajectory diversification.
To address the limitations of PDM in complex scenarios, we introduced the SPDM approach in Section \ref{subsec:spdm}, designed to generate more sophisticated driving maneuvers. As shown in Table \ref{tab:proposals_spdm_pdm}, the inclusion of SPDM generated proposals results in a $50\%$ improvement in driving performance on the interPlan benchmark compared to the original PDM proposal set. In contrast, increasing the number of PDM proposals alone does not yield additional performance gains and actually leads to a slight decrease in performance. A similar trend is observed for SPDM on the Val14 benchmark.
Finally, we show that our approach also is compute-efficient. Figure \ref{fig:runtime} presents a runtime comparison between PDM-Closed, Diffusion-ES, and our proposed method using different numbers of proposals. We include only Diffusion-ES in this comparison, as it represents the best-performing learning-based baseline on interPlan. Our method shows a slightly higher computational cost compared to PDM-Closed when using a small number of proposals, mainly because of the additional processing needed to incorporate neighboring lanes. However, as the number of proposals increases, the runtime becomes comparable to that of PDM-Closed. This is because PDM-Closed adapts proposals to predictions, while our method generates proposals independently, making the proposal generation process more efficient. In contrast, Diffusion-ES demonstrates significantly lower efficiency, with a large increase in runtime compared to SPDM and PDM-Closed.

%\begin{figure}[t]
%\centering
%\begin{center}
%\includegraphics[width=.8\columnwidth]{figures/fig7.pdf}
%\end{center}
%\caption{Runtime analysis across varying numbers of proposals, averaged over 10 distinct scenarios.}
%\label{fig:runtime}
%\end{figure}

% ----------------------------- Conclusion ----------------------------
\section{Conclusion}
\label{sec:conclusion}
In this work, we demonstrated that current Integrated Prediction and Planning (IPP) approaches face substantial challenges in effectively leveraging predictions, with many existing planning methods further constrained by their limited ability to generate high-quality trajectories. In contrast, our proposed proposal generation method delivers a notable $50\%$ performance improvement in out-of-distribution scenarios. These findings highlight a critical need to rethink how prediction is integrated into planning and to advance proposal generation as a central component of future IPP systems.

% ----------------------------- Bib ------------------------------
\bibliographystyle{IEEEtran}
\bibliography{main}

% ----------------------------- Appendix ------------------------------
%\addtolength{\textheight}{-12cm}   
%\clearpage
%\input{sections/appendix}

\end{document}